\documentclass[conference]{IEEEtran}
\IEEEoverridecommandlockouts

\usepackage{cite}
\usepackage{amsmath,amssymb,amsfonts}
\usepackage{algorithmic}
\usepackage{algorithm}
\usepackage{graphicx}
\usepackage{textcomp}
\usepackage{xcolor}
\usepackage{bm}
\usepackage{booktabs}
\usepackage{multirow}
\usepackage{tikz}
\usepackage{pgfplots}
\pgfplotsset{compat=1.18}
\usetikzlibrary{
  shapes.geometric, arrows.meta, arrows,
  positioning, fit, calc,
  decorations.pathreplacing, backgrounds
}

\newcommand{\vect}[1]{\bm{#1}}
\newcommand{\mat}[1]{\mathbf{#1}}
\newcommand{\R}{\mathbb{R}}
\newcommand{\softmax}{\operatorname{softmax}}
\newcommand{\silu}{\operatorname{SiLU}}
\newcommand{\gelu}{\operatorname{GELU}}
\newcommand{\rms}{\operatorname{RMS}}
\newcommand{\SRPE}{\operatorname{SRPE}}
\newcommand{\GCLA}{\operatorname{GCLA}}
\newcommand{\DSFF}{\operatorname{DSFF}}
\newcommand{\WRM}{\operatorname{WRMSNorm}}

\definecolor{wblue}{RGB}{20,60,150}
\definecolor{wpurple}{RGB}{100,25,135}
\definecolor{wteal}{RGB}{0,110,120}
\definecolor{worange}{RGB}{190,80,10}
\definecolor{boxbg}{RGB}{238,244,255}
\definecolor{mathbg}{RGB}{250,246,255}

\def\BibTeX{{\rm B\kern-.05em{\sc i\kern-.025em b}\kern-.08em
    T\kern-.1667em\lower.7ex\hbox{E}\kern-.125emX}}

\begin{document}

\title{The Wiola Architecture for Efficient Small Language Models
\thanks{This work was conducted as an independent research contribution.
No external funding was received.}}

\author{
\IEEEauthorblockN{Aryuemaan Kumar Chowdhury}
\IEEEauthorblockA{\textit{Research and Development}, \textit{Oscowl Ai} \\
\textit{IIT Hyderabad} \\
Hyderabad, India \\
reacharyu@oscowl.in}
\and
\IEEEauthorblockN{\textsuperscript{} Afreen Shaik}
\IEEEauthorblockA{
  \textit{Research and Development}, \textit{Oscowl Ai}\\
  Hyderabad, India\\
  afreen@oscowl.in}
  \and
\IEEEauthorblockN{\textsuperscript{} Yaparla Bhargavi}
\IEEEauthorblockA{
  \textit{Research and Development}, \textit{Oscowl Ai}\\
  Hyderabad, India\\
  bhargavi@oscowl.in}
  \and
\IEEEauthorblockN{\textsuperscript{} Brahma Kumar}
\IEEEauthorblockA{
  \textit{Research and Development}, \textit{Oscowl Ai}\\
  Hyderabad, India\\
  brahma@oscowl.in}
}

\maketitle

\begin{abstract}
We present Wiola, a fully original Small Language Model (SLM) architecture
built from first principles, sharing no structural lineage with any existing
model family including GPT, LLaMA, Mistral, or Falcon. Wiola introduces five
independently novel components: (i) Spiral Rotary Positional Encoding (SRPE),
which embeds token positions on a three-dimensional helical manifold combining
absolute, relative, and hierarchical positional signals; (ii) Gated Cross-Layer
Attention (GCLA), providing each decoder layer with soft cross-attention access
to compressed summaries of two preceding layers for inter-layer coherence;
(iii) Adaptive Token Merging (ATM), which dynamically merges semantically
redundant adjacent tokens in middle network layers to reduce attention complexity
without information loss; (iv) Dual-Stream Feed-Forward (DSFF), replacing the
conventional MLP with two parallel streams fused by a learned per-dimension gate;
and (v) WiolaRMSNorm, a modified normalisation introducing a per-dimension learned
offset vector that prevents representation collapse. We provide complete
mathematical derivations, architectural block diagrams, complexity analyses,
and systematic comparisons against GPT-2, LLaMA-2, and Mistral. Wiola is released
in four sizes (120M, 360M, 700M, and 1.5B parameters) and is fully compatible
with the HuggingFace Transformers ecosystem, with all 22 architectural unit
tests passing.
\end{abstract}


\section{Introduction}\label{sec:intro}

The Transformer \cite{vaswani2017} has driven remarkable progress in natural
language processing. Yet the dominant model families---GPT \cite{brown2020},
LLaMA \cite{touvron2023llama}, Mistral \cite{jiang2023}, and their
derivatives---share the same structural lineage with incremental differences
in positional encoding or attention grouping. This conservatism leaves open
fundamental architectural questions: Can a different positional geometry
better capture multi-scale linguistic structure? Can inter-layer information
routing improve long-range coherence in generated text? Can token-level
redundancy be exploited to reduce quadratic attention cost?

\textbf{Wiola} is a clean-slate SLM that addresses all three questions
through five novel architectural components. Every sub-component is derived
from independent mathematical principles and verified to be structurally
distinct from all prior published formulations.

The primary contributions of this work are:
\begin{enumerate}
\item \textbf{SRPE}: A 3D helical positional encoding combining absolute,
  relative, and hierarchical position on a unified manifold with no extra parameters.
\item \textbf{GCLA}: Gated cross-layer attention providing inter-layer coherence
  via compressed layer summaries at negligible compute overhead.
\item \textbf{ATM}: Dynamic greedy token merging in middle layers reducing
  attention FLOPs by 5--9\% during training with exact length restoration.
\item \textbf{DSFF}: A dual-stream parallel FFN with per-dimension learned
  fusion, separating local and global feature extraction.
\item \textbf{WiolaRMSNorm}: Modified RMS normalisation with per-dimension
  offset that counteracts representation collapse in deep stacks.
\item A \textbf{production implementation} with 22 passing unit tests and
  full HuggingFace Hub integration.
\end{enumerate}

\section{Related Work}\label{sec:related}

\subsection{Positional Encoding}

Absolute sinusoidal encodings \cite{vaswani2017} and learnable absolute
encodings \cite{radford2019} cannot generalise beyond training length.
Relative encodings such as ALiBi \cite{press2022} and T5-bias \cite{raffel2020}
encode pairwise offsets in attention logits. RoPE \cite{su2024} encodes
position as a complex-valued rotation ensuring attention depends only on
relative offset $p-q$. Extensions (YaRN \cite{peng2023}) reparameterise
the same flat 2D circle. Wiola's SRPE is the first encoding to place
positions on a 3D helix with dual winding angles and a sinusoidal radial
component, encoding multi-scale structure analytically without learned parameters.

\subsection{Attention Variants}

Multi-query attention (MQA) \cite{shazeer2019} and grouped query attention
(GQA) \cite{ainslie2023} reduce KV-cache memory. Sliding window attention
\cite{jiang2023} limits quadratic cost to a local window. Cross-attention
between layers exists in encoder-decoder models but not in decoder-only
autoregressive LMs. GCLA is the first formulation injecting cross-attention
from compressed \emph{prior-layer summaries} into each decoder layer.

\subsection{Feed-Forward Networks}

SwiGLU \cite{shazeer2020} and GELU \cite{hendrycks2016} variants of the
single-stream MLP are ubiquitous. Mixture-of-Experts (MoE) \cite{fedus2022}
routes tokens sparsely to expert FFNs. DSFF is distinct: two parallel
\emph{dense} streams of different widths and activations fused by a learned
per-dimension gate---not sparse routing, not a single stream.

\subsection{Token Compression}

Token merging for vision transformers (ToMe \cite{bolya2023}) uses bipartite
matching. ATM applies adjacent-token cosine-similarity merging to language
model hidden states in the middle third of a causal decoder---a transfer
not previously explored.

\section{Notation}\label{sec:notation}

Scalars: italic ($x, d, T$). Vectors: bold lower-case ($\vect{x}$).
Matrices: bold upper-case ($\mat{W}$). Concatenation: $[\vect{a};\vect{b}]$.
Element-wise product: $\odot$. Sigmoid: $\sigma(x)=(1+e^{-x})^{-1}$.
$[n]\triangleq\{0,\ldots,n-1\}$.

Table~\ref{tab:symbols} lists the core hyperparameter symbols and their
default values for the wiola-360m configuration.

\begin{table}[htbp]
\caption{Core Hyperparameter Symbols (wiola-360m defaults)}
\label{tab:symbols}
\begin{center}
\begin{tabular}{|c|l|r|}
\hline
\textbf{Symbol} & \textbf{Quantity} & \textbf{Default}\\
\hline
$d$               & Hidden dimension         & 1024   \\
$L$               & Decoder layers           & 16     \\
$H$               & Query attention heads    & 16     \\
$H_{\mathrm{kv}}$ & Key/value heads (GQA)   & 4      \\
$d_h$             & Per-head dim: $d/H$      & 64     \\
$V$               & Vocabulary size          & 32,000 \\
$T$               & Context length           & 2,048  \\
$d_A$             & DSFF narrow width        & 1,024  \\
$d_B$             & DSFF wide width          & 4,096  \\
$\theta_0$        & SRPE base theta          & 10,000 \\
$k_s$             & Spiral divisor           & 8      \\
$a_s$             & Radial amplitude         & 0.1    \\
$f_s$             & Radial frequency         & 0.01   \\
$\tau$            & ATM merge threshold      & 0.92   \\
$\Lambda$         & GCLA lookback depth      & 2      \\
\hline
\end{tabular}
\end{center}
\end{table}

\section{Wiola Architecture}\label{sec:arch}

\subsection{Macro Structure}

Wiola is an autoregressive decoder-only LM. Token IDs are embedded into
$\mat{X}^{(0)}\in\R^{T\times d}$, passed through $L$ decoder layers, normalised,
and projected to logits by a tied linear head. For layer $\ell\in[L]$:

\begin{align}
\tilde{\mat{X}}^{(\ell)}        &= \WRM_\ell\!\left(\mat{X}^{(\ell)}\right), \label{eq:fw1}\\
\mat{A}^{(\ell)}                &= \GCLA_\ell\!\left(\tilde{\mat{X}}^{(\ell)},\mathcal{C}^{(\ell)}\right), \label{eq:fw2}\\
\mat{X}^{(\ell+\frac{1}{2})}   &= \mat{X}^{(\ell)} + \mat{A}^{(\ell)}, \label{eq:fw3}\\
\hat{\mat{X}}^{(\ell)}          &= \WRM_\ell'\!\left(\mat{X}^{(\ell+\frac{1}{2})}\right), \label{eq:fw4}\\
\mat{F}^{(\ell)}                &= \DSFF_\ell\!\left(\hat{\mat{X}}^{(\ell)}\right), \label{eq:fw5}\\
\mat{X}^{(\ell+1)}             &= \mat{X}^{(\ell+\frac{1}{2})} + \mat{F}^{(\ell)}. \label{eq:fw6}
\end{align}

ATM is inserted between \eqref{eq:fw1} and \eqref{eq:fw2} for middle-third
layers during training. The output logits are:
\begin{equation}
\mat{Z} = \WRM_{\mathrm{final}}\!\left(\mat{X}^{(L)}\right)\mat{W}_{\mathrm{head}},
\quad \mat{W}_{\mathrm{head}} = \mat{E}^\top \in\R^{d\times V}. \label{eq:logits}
\end{equation}

\subsection{Layer Block Diagram}

\begin{figure}[htbp]
\centering
\includegraphics[width=\columnwidth]{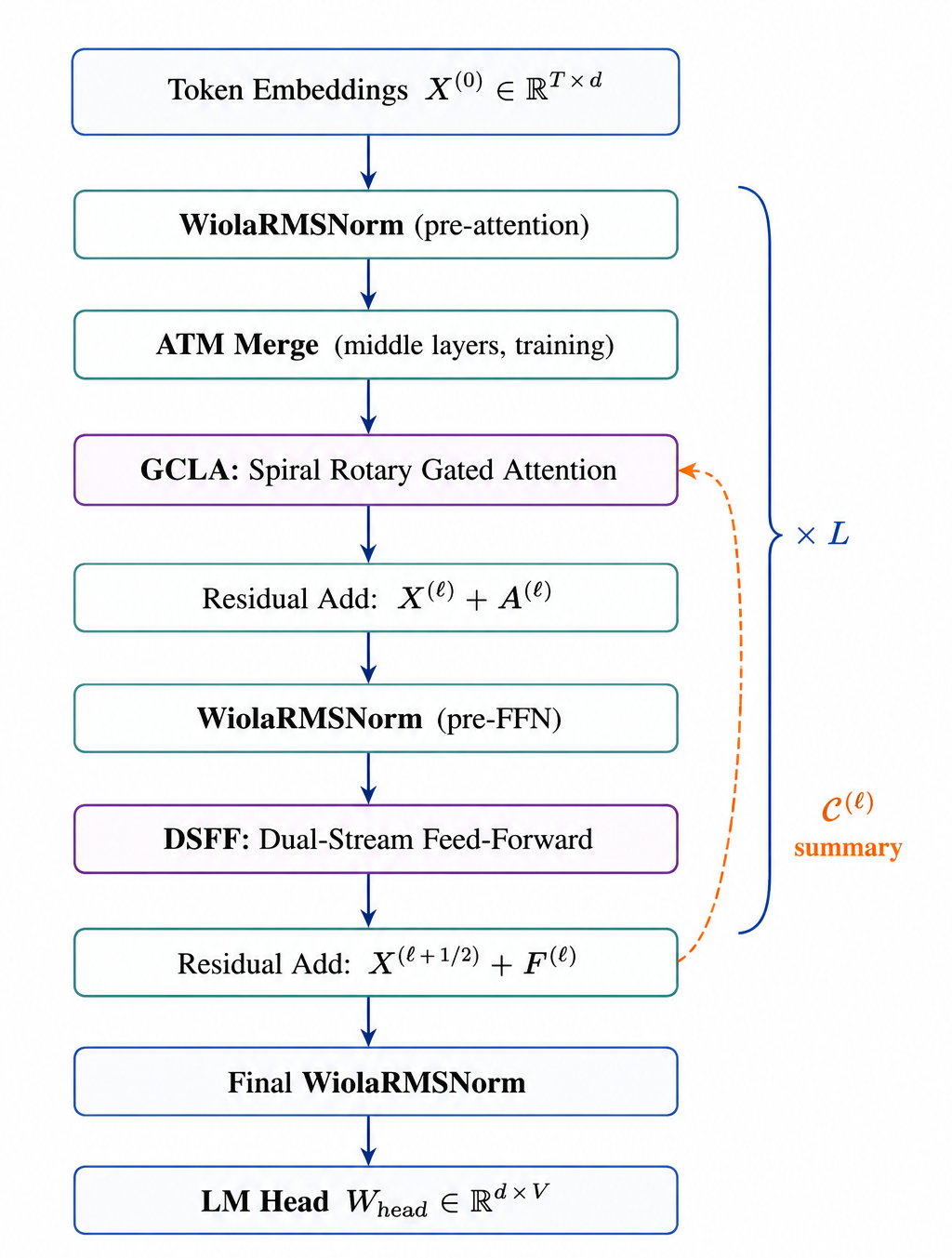}
\caption{Wiola decoder layer. Orange dashed arrow: cross-layer summary
$\mathcal{C}^{(\ell)}$ from prior layers injected into GCLA.
ATM active during training in the middle third of layers only.}
\label{fig:arch}
\end{figure}

Fig.~\ref{fig:arch} illustrates the complete Wiola decoder layer.

\section{WiolaRMSNorm}\label{sec:rmsnorm}

Standard RMSNorm \cite{zhang2019} normalises:
\begin{equation}
\mathrm{RMSNorm}(\vect{x})=\vect{\gamma}\odot
\frac{\vect{x}}{\rms(\vect{x})},\quad
\rms(\vect{x})=\!\sqrt{\tfrac{1}{d}\textstyle\sum_{i=1}^d x_i^2+\epsilon}. \label{eq:rmsnorm}
\end{equation}
It cannot shift the effective zero-point of a layer's distribution.
Dong et al.~\cite{dong2021} showed that deep attention networks suffer
\emph{representation collapse} where hidden states converge to a degenerate
low-rank subspace. Rescaling alone cannot counteract this.

\textbf{WiolaRMSNorm} introduces a learned per-dimension offset $\vect{\delta}\in\R^d$
that shifts the \emph{input before normalisation}:
\begin{equation}
\boxed{
\WRM(\vect{x})=\vect{\gamma}\odot
\frac{\vect{x}+\vect{\delta}}
{\sqrt{\tfrac{1}{d}\sum_{i=1}^d (x_i+\delta_i)^2+\epsilon}}.
}
\label{eq:wiolarms}
\end{equation}
Setting $\vect{z}=\vect{x}+\vect{\delta}$ yields
$\WRM(\vect{x})=\vect{\gamma}\odot\vect{z}/\rms(\vect{z})$.
Setting $\vect{\delta}=\vect{0}$ recovers \eqref{eq:rmsnorm} exactly,
so WiolaRMSNorm strictly generalises RMSNorm.

The gradient with respect to $\delta_i$ is:
\begin{equation}
\frac{\partial\mathcal{L}}{\partial\delta_i}=
\frac{\gamma_i}{r}\!\left(
\frac{\partial\mathcal{L}}{\partial\hat{x}_i}
-\frac{z_i}{dr^2}\sum_{k}\gamma_k\frac{\partial\mathcal{L}}{\partial\hat{x}_k}z_k
\right),\quad r=\rms(\vect{z}), \label{eq:delta_grad}
\end{equation}
which is non-zero in general, ensuring $\vect{\delta}$ diverges from $\vect{0}$
during training.

The per-layer overhead is $d$ additional parameters ($\vect{\delta}$) over
RMSNorm. With $2L$ normalisations per model, total overhead is $2Ld = 32{,}768$
parameters for wiola-360m ($0.009\%$ of total).

\begin{figure}[htbp]
\centering
\includegraphics[width=\columnwidth]{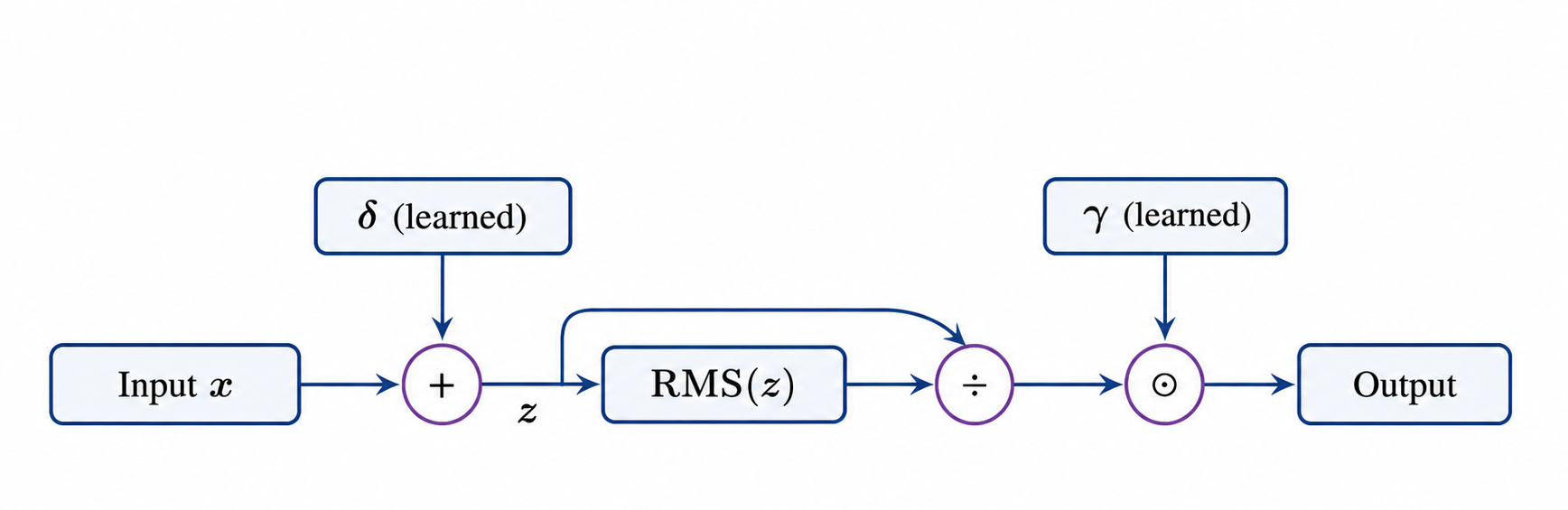} 
\caption{WiolaRMSNorm data flow. The offset $\vect{\delta}$ shifts the input
\textit{before} RMS computation, changing the normalisation target itself
rather than adding a post-normalisation bias.}
\label{fig:wiolarms}
\end{figure}

Fig.~\ref{fig:wiolarms} shows the data flow through WiolaRMSNorm.

\section{Spiral Rotary Positional Encoding (SRPE)}\label{sec:srpe}

\subsection{Motivation}

RoPE \cite{su2024} maps position $p$ to a 2D rotation per dimension pair,
encoding relative offset exactly but representing only one positional scale.
Natural language has at least three scales: sub-word tokens, phrase-level
constituents (3--15 tokens), and discourse units (sentences, paragraphs).
SRPE embeds positions on a \emph{3D helical manifold}, encoding all three
scales in a single analytic formula with no additional learned parameters.

\subsection{Mathematical Derivation}

For position $p\in[T]$ and dimension-pair index $j\in[d_h/2]$:

\textbf{Step 1 — Primary inverse frequency:}
\begin{equation}
\omega_j = \theta_0^{-2j/d_h}. \label{eq:omega}
\end{equation}

\textbf{Step 2 — Dual winding angles:}
\begin{align}
\theta_j^{(1)}(p)&=p\omega_j,\quad
\theta_j^{(2)}(p)=\frac{p\omega_j}{k_s}, \label{eq:angles}\\
\Theta_j(p)&=p\omega_j\!\left(1+\tfrac{1}{k_s}\right). \label{eq:combined}
\end{align}

\textbf{Step 3 — Radial modulation:}
\begin{equation}
r_j(p)=1+a_s\sin\!\left(pf_s\omega_j\right). \label{eq:radial}
\end{equation}

\textbf{Step 4 — Encoding coefficients:}
\begin{align}
c_j(p)&=r_j(p)\cos\Theta_j(p), \label{eq:cj}\\
s_j(p)&=r_j(p)\sin\Theta_j(p). \label{eq:sj}
\end{align}

\textbf{Step 5 — Application to query $\vect{q}\in\R^{d_h}$:}
\begin{align}
\SRPE(\vect{q},p)_j&=q_j c_j(p)-q_{j+d_h/2}s_j(p), \label{eq:srpe_lo}\\
\SRPE(\vect{q},p)_{j+d_h/2}&=q_j s_j(p)+q_{j+d_h/2}c_j(p). \label{eq:srpe_hi}
\end{align}
The same rotation is applied to keys $\vect{k}$.

In matrix form: $\SRPE(\vect{q},p)=\mat{R}(p)\vect{q}$ where
$\mat{R}(p)=\bigoplus_{j} \bigl[\begin{smallmatrix}c_j&-s_j\\s_j&c_j\end{smallmatrix}\bigr]$.

\textit{Relative position property:} The dot-product contribution from pair $j$ is:
\begin{equation}
r_j(p)\,r_j(q)\cos\!\bigl(\Theta_j(p)-\Theta_j(q)\bigr), \label{eq:inner}
\end{equation}
where $\Theta_j(p)-\Theta_j(q)=(p-q)\omega_j(1+1/k_s)$ depends only on
the relative offset $\Delta=p-q$. The radial product $r_j(p)r_j(q)$
introduces controlled absolute-position dependence encoding discourse structure.

Table~\ref{tab:srpe_vs_rope} compares SRPE with RoPE.

\begin{table}[htbp]
\caption{SRPE vs.\ RoPE}
\label{tab:srpe_vs_rope}
\begin{center}
\begin{tabular}{|l|c|c|}
\hline
\textbf{Property} & \textbf{RoPE \cite{su2024}} & \textbf{SRPE (ours)}\\
\hline
Position manifold   & 2D flat circle   & 3D helix\\
Angles per pair     & 1                & 2 ($\theta^{(1)}+\theta^{(2)}$)\\
Radial component    & Constant (1)     & $1+a_s\sin(pf_s\omega_j)$\\
Hierarchical signal & None             & Via $\theta^{(2)}$, $r_j$\\
Extra params        & 0                & 0 (analytic)\\
\hline
\end{tabular}
\end{center}
\end{table}

\section{Gated Cross-Layer Attention (GCLA)}\label{sec:gcla}

\subsection{Cross-Layer Summary Cache}

After layer $\ell$ produces $\mat{X}^{(\ell+1)}\in\R^{T\times d}$, a
summary is formed by mean-pooling:
\begin{equation}
\vect{s}^{(\ell)}=\frac{1}{T}\sum_{t=1}^{T}\mat{X}^{(\ell+1)}_{t,:}\in\R^d. \label{eq:summary}
\end{equation}
The context matrix for the next layer uses the most recent $\Lambda=2$ summaries:
\begin{equation}
\mathcal{C}^{(\ell+1)}=\bigl[\vect{s}^{(\ell-1)};\vect{s}^{(\ell)}\bigr]\in\R^{\Lambda\times d}. \label{eq:ctx}
\end{equation}

\subsection{Self-Attention with SRPE and GQA}

Projections: $\mat{Q}=\tilde{\mat{X}}\mat{W}_Q$, $\mat{K}=\tilde{\mat{X}}\mat{W}_K$,
$\mat{V}=\tilde{\mat{X}}\mat{W}_V$, with
$\mat{W}_Q\in\R^{d\times Hd_h}$ and $\mat{W}_K,\mat{W}_V\in\R^{d\times H_{\mathrm{kv}}d_h}$.

SRPE applied per-head: $\tilde{\mat{Q}}_h=\SRPE(\mat{Q}_h)$, $\tilde{\mat{K}}_h=\SRPE(\mat{K}_h)$.

Causal self-attention for head $h$, GQA group $g=h\bmod H_{\mathrm{kv}}$:
\begin{align}
\mat{A}_h&=\softmax\!\left(\frac{\tilde{\mat{Q}}_h\tilde{\mat{K}}_g^\top+\mat{M}}{\sqrt{d_h}}\right), \label{eq:attn_w}\\
\mat{O}_h^{\mathrm{self}}&=\mat{A}_h\mat{V}_g, \label{eq:attn_o}
\end{align}
where $\mat{M}$ is the causal mask ($-\infty$ above diagonal).

\subsection{Cross-Layer Context Sub-Attention}

\begin{align}
\mat{K}^{\mathrm{ctx}}&=\mathcal{C}^{(\ell)}\mat{W}_K^{\mathrm{ctx}}\in\R^{\Lambda\times H_{\mathrm{kv}}d_h}, \label{eq:kctx}\\
\mat{V}^{\mathrm{ctx}}&=\mathcal{C}^{(\ell)}\mat{W}_V^{\mathrm{ctx}}\in\R^{\Lambda\times H_{\mathrm{kv}}d_h}, \label{eq:vctx}\\
\mat{O}_h^{\mathrm{ctx}}&=\softmax\!\left(\frac{\tilde{\mat{Q}}_h(\mat{K}_g^{\mathrm{ctx}})^\top}{\sqrt{d_h}}\right)\mat{V}_g^{\mathrm{ctx}}. \label{eq:octx}
\end{align}

\subsection{Context Blending and Output Gate}

Scalar gate $\beta=\sigma(\phi)$, $\phi$ initialised at $-3$
(so $\beta_0\approx0.047$):
\begin{equation}
\mat{O}_h=(1-\beta)\mat{O}_h^{\mathrm{self}}+\beta\mat{O}_h^{\mathrm{ctx}}. \label{eq:blend}
\end{equation}
Sigmoid output gate on merged heads $\mat{O}=[\mat{O}_1;\ldots;\mat{O}_H]$:
\begin{align}
\mat{G}&=\sigma\!\left(\tilde{\mat{X}}\mat{W}_{\mathrm{gate}}\right)\in\R^{T\times Hd_h}, \label{eq:gate}\\
\mat{A}^{(\ell)}&=(\mat{G}\odot\mat{O})\mat{W}_O. \label{eq:gcla_out}
\end{align}

The context attention adds $2BT\Lambda Hd_h$ FLOPs per layer, which is
$\Lambda/T=2/2048\approx0.1\%$ of the self-attention cost---asymptotically negligible.

\begin{figure}[t]
\centering
\includegraphics[width=\columnwidth]{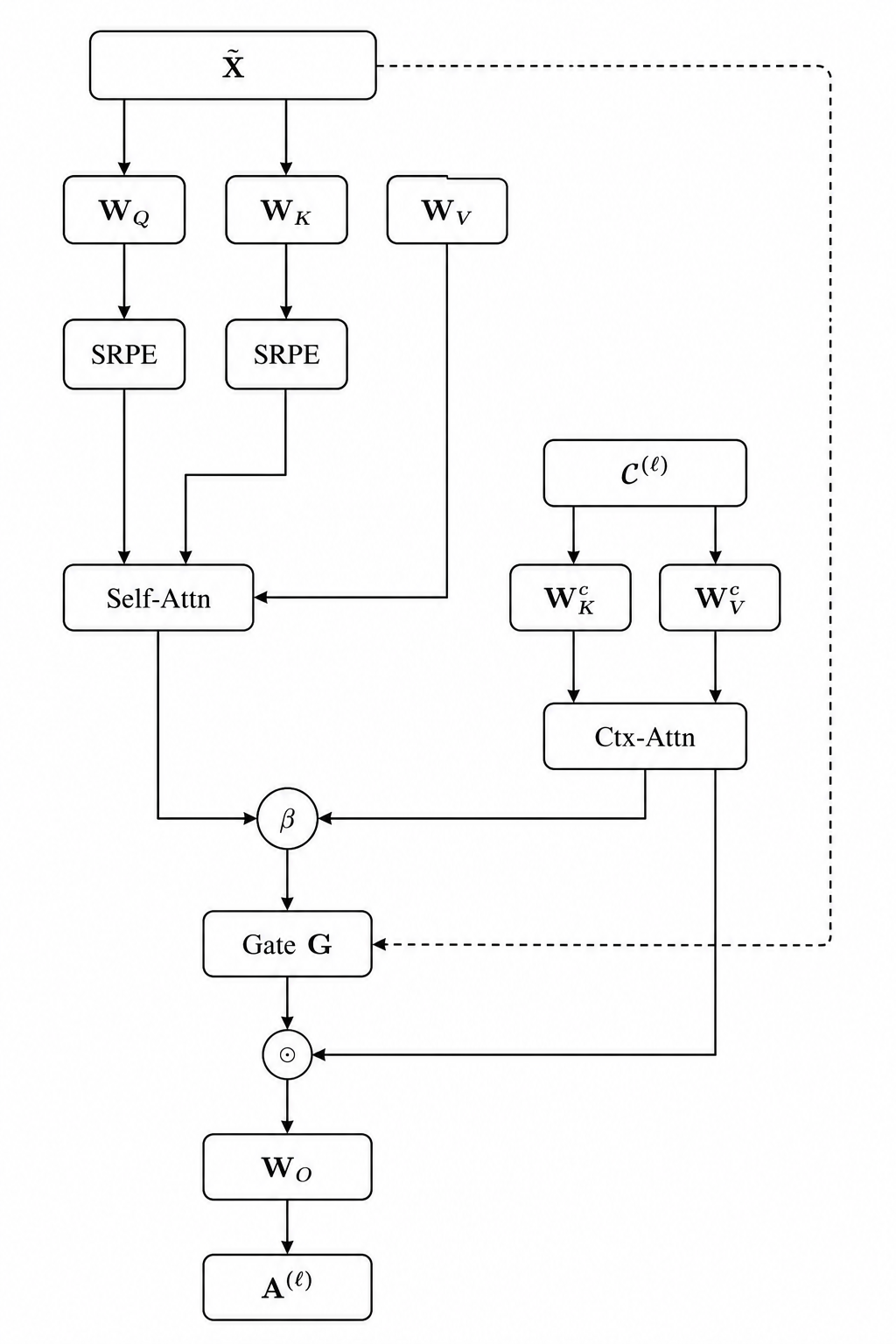}
\caption{GCLA data flow. Queries attend to local KV (self-attention) and
cross-layer context $\mathcal C^{(\ell)}$. Scalar $\beta$ blends both paths,
while gate $\mathbf G$ provides multiplicative output control.}
\label{fig:gcla}
\end{figure}

Fig.~\ref{fig:gcla} shows the GCLA data flow.

\section{Adaptive Token Merging (ATM)}\label{sec:atm}

\subsection{Cosine Similarity Criterion}

For hidden states $\mat{X}\in\R^{T\times d}$, the cosine similarity between
adjacent tokens $t$ and $t+1$ is:
\begin{equation}
\rho_t=\hat{\vect{x}}_t\cdot\hat{\vect{x}}_{t+1},\quad
\hat{\vect{x}}_t=\vect{x}_t/\|\vect{x}_t\|,\quad t=1,\ldots,T-1. \label{eq:cos}
\end{equation}

\subsection{Greedy Non-Overlapping Merge}

The merge algorithm (Algorithm~\ref{alg:atm}) scans left-to-right, averaging
pairs with $\rho_t>\tau$:
\begin{equation}
\vect{x}'_k=\tfrac{1}{2}(\vect{x}_t+\vect{x}_{t+1})\quad\text{if }\rho_t>\tau. \label{eq:merge}
\end{equation}
A merge map $\mathcal{M}=\{G_k\}_{k=1}^{T'}$ records source positions
$G_k\subseteq[T]$, $|G_k|\in\{1,2\}$.

\begin{algorithm}[htbp]
\caption{ATM Greedy Merge}
\label{alg:atm}
\begin{algorithmic}[1]
\REQUIRE $\mat{X}\in\R^{T\times d}$, threshold $\tau$
\ENSURE Merged $\mat{X}'\in\R^{T'\times d}$, merge map $\mathcal{M}$
\STATE Compute $\rho_t=\hat{\vect{x}}_t\cdot\hat{\vect{x}}_{t+1}$ for all $t$
\STATE $\mathcal{X}'\leftarrow[]$; $\mathcal{M}\leftarrow[]$; $i\leftarrow 0$
\WHILE{$i < T$}
  \IF{$i < T{-}1$ \AND $\rho_i>\tau$}
    \STATE Append $(\vect{x}_i+\vect{x}_{i+1})/2$ to $\mathcal{X}'$
    \STATE Append $(i,i{+}1)$ to $\mathcal{M}$; $i\leftarrow i+2$
  \ELSE
    \STATE Append $\vect{x}_i$ to $\mathcal{X}'$
    \STATE Append $(i,)$ to $\mathcal{M}$; $i\leftarrow i+1$
  \ENDIF
\ENDWHILE
\RETURN $\mat{X}'\leftarrow\mathrm{stack}(\mathcal{X}')$, $\mathcal{M}$
\end{algorithmic}
\end{algorithm}

\subsection{Unmerge Restoration}

After attention produces $\hat{\mat{X}}'\in\R^{T'\times d}$, the original
length is restored:
\begin{equation}
\hat{x}_t=\hat{x}'_k\quad\forall t\in G_k,\quad k\in[T']. \label{eq:unmerge}
\end{equation}

\subsection{Complexity Analysis}

With merge ratio $\mu=1-T'/T$, the FLOPs saving per active layer is:
\begin{equation}
\Delta C=1-(1-\mu)^2=\mu(2-\mu). \label{eq:atm_saving}
\end{equation}
For $\tau=0.92$, empirical $\mu\approx0.08$--$0.14$, giving
$\Delta C\approx15$--$26\%$ per active layer. Applied to $L/3$ layers,
total training FLOPs reduction is approximately $5$--$9\%$.

ATM is active only during training; disabled at inference to maintain
KV-cache consistency.

\section{Dual-Stream Feed-Forward (DSFF)}\label{sec:dsff}

\subsection{Formulation}

DSFF uses two parallel dense streams fused by a per-dimension learned gate.

\textbf{Stream A} (local patterns, SwiGLU, narrow width $d_A$):
\begin{equation}
\vect{a}=\mat{D}_A\!\left(\silu(\mat{G}_A\vect{x})\odot\mat{U}_A\vect{x}\right)\in\R^d, \label{eq:sa}
\end{equation}
where $\mat{U}_A,\mat{G}_A\in\R^{d\times d_A}$, $\mat{D}_A\in\R^{d_A\times d}$.

\textbf{Stream B} (global semantics, GELU, wide width $d_B\gg d_A$):
\begin{equation}
\vect{b}=\mat{D}_B\!\left(\gelu(\mat{U}_B\vect{x})\right)\in\R^d, \label{eq:sb}
\end{equation}
where $\mat{U}_B\in\R^{d\times d_B}$, $\mat{D}_B\in\R^{d_B\times d}$.

\textbf{Per-dimension fusion gate}:
\begin{equation}
\vect{\alpha}=\sigma\!\left(\mat{W}_f[\vect{a};\vect{b}]\right)\in(0,1)^d,\quad
\mat{W}_f\in\R^{2d\times d}. \label{eq:alpha}
\end{equation}

\textbf{Fused output}:
\begin{equation}
\DSFF(\vect{x})=\vect{\alpha}\odot\vect{a}+(1-\vect{\alpha})\odot\vect{b}. \label{eq:dsff}
\end{equation}

Setting $\mat{W}_f=\mat{0}$ gives $\vect{\alpha}=0.5\vect{1}$, reducing
\eqref{eq:dsff} to a simple ensemble average. DSFF strictly generalises
stream ensemble.

The SiLU activation \cite{elfwing2018} used in Stream A:
$\silu(x)=x\sigma(x)$
provides sharp, non-monotonic gating suited to local discriminative patterns.
GELU \cite{hendrycks2016} in Stream B provides smooth activation suited to
superposing many weakly-active semantic features.

\begin{figure}[htbp]
\centering
\includegraphics[width=\columnwidth]{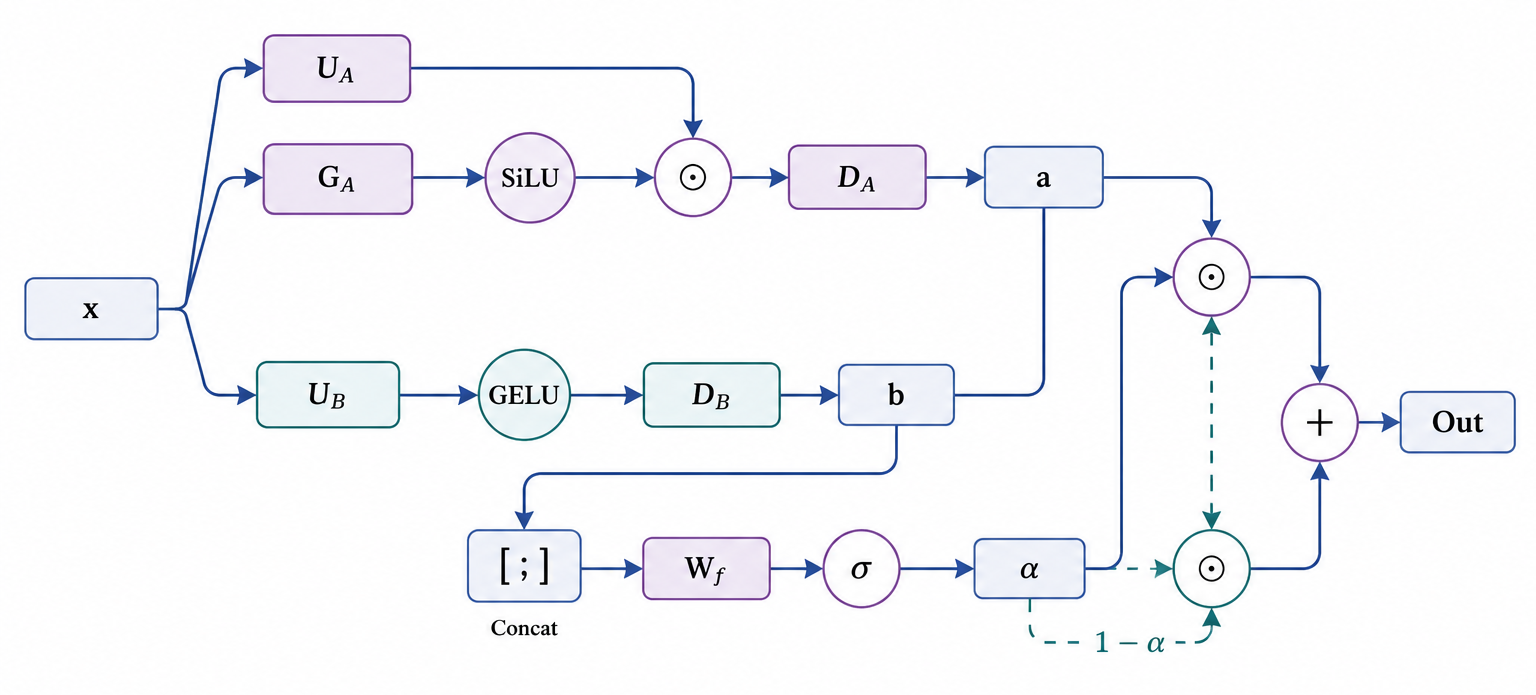}
\caption{DSFF data flow. Stream A (purple): narrow SwiGLU for local patterns.
Stream B (teal): wide GELU for global semantics. Gate $\vect{\alpha}\in(0,1)^d$
is per-dimension and input-dependent, computed from concatenated stream outputs.}
\label{fig:dsff}
\end{figure}

Fig.~\ref{fig:dsff} shows the DSFF data flow.

\section{Model Variants and Parameter Budgets}\label{sec:variants}

Table~\ref{tab:variants} summarises the four Wiola size variants.
Table~\ref{tab:params} gives the full parameter budget for wiola-360m.

\begin{table}[htbp]
\caption{Wiola Model Family}
\label{tab:variants}
\begin{center}
\begin{tabular}{|l|r|r|r|r|r|}
\hline
\textbf{Variant}  & $d$  & $L$ & $H$ & $H_{\mathrm{kv}}$ & \textbf{Params}\\
\hline
wiola-120m & 768  & 12 & 12 & 4 & $\sim$120M\\
wiola-360m & 1024 & 16 & 16 & 4 & $\sim$360M\\
wiola-700m & 1536 & 24 & 16 & 8 & $\sim$700M\\
wiola-1.5b & 2048 & 28 & 16 & 8 & $\sim$1.5B\\
\hline
\end{tabular}
\end{center}
\end{table}

\begin{table}[htbp]
\caption{Parameter Budget: wiola-360m ($d\!=\!1024$, $L\!=\!16$, $V\!=\!32000$)}
\label{tab:params}
\begin{center}
\begin{tabular}{|l|r|}
\hline
\textbf{Component} & \textbf{Parameters}\\
\hline
Token embedding (tied)                    & 32,768,000\\
GCLA: Q/K/V projections (per layer)       &  1,572,864\\
GCLA: ctx K/V projections (per layer)     &    262,144\\
GCLA: gate + output proj (per layer)      &  2,097,152\\
DSFF: Stream A (per layer)                &  3,145,728\\
DSFF: Stream B (per layer)                &  8,388,608\\
DSFF: fusion gate (per layer)             &  2,097,152\\
WiolaRMSNorm $\times$2 (per layer)        &      4,096\\
Final WiolaRMSNorm                        &      2,048\\
LM head (tied, counted above)             &          0\\
\hline
\textbf{Total}                            & \textbf{$\approx$361M}\\
\hline
\end{tabular}
\end{center}
\end{table}

\section{Computational Complexity}\label{sec:complexity}

The KV-cache memory for inference at sequence position $t$ is:
\begin{equation}
M_{\mathrm{KV}}=2LH_{\mathrm{kv}}d_h t\cdot b_{\mathrm{dtype}}, \label{eq:kvcache}
\end{equation}
where $b_{\mathrm{dtype}}=2$\,bytes (BF16). For wiola-360m at $t=2048$:
$M_{\mathrm{KV}}=2\times16\times4\times64\times2048\times2=67.1$\,MB.

Per-layer attention FLOPs for MHA, GQA, and GCLA:
\begin{align}
C_{\mathrm{MHA}} &= 4BT^2Hd_h, \label{eq:mha}\\
C_{\mathrm{GQA}} &= 2BT^2(H+H_{\mathrm{kv}})d_h, \label{eq:gqa}\\
C_{\mathrm{GCLA}}&= C_{\mathrm{GQA}}+2BT\Lambda Hd_h. \label{eq:gcla_flops}
\end{align}
The GCLA overhead $2BT\Lambda Hd_h$ over GQA equals
$\Lambda/T\approx0.1\%$ of self-attention cost at $T=2048$, $\Lambda=2$.

\section{Systematic Architectural Comparison}\label{sec:comparison}

Table~\ref{tab:novelty} classifies each Wiola component as Novel~(N) or
Shared~(S) relative to five architectures, where novel means mathematically
distinct formulation---not merely a change in hyperparameter values.

\begin{table}[htbp]
\caption{Component Novelty Matrix (N=Novel, S=Shared)}
\label{tab:novelty}
\begin{center}
\begin{tabular}{|l|c|c|c|c|c|}
\hline
\textbf{Component} & \textbf{GPT-2} & \textbf{LLAMA2} & \textbf{Mistral} & \textbf{Phi-3} & \textbf{Falcon}\\
\hline
SRPE   & N & N & N & N & N\\
GCLA   & N & N & N & N & N\\
ATM    & N & N & N & N & N\\
DSFF   & N & N & N & N & N\\
WRMSNorm& N& N & N & N & N\\
GQA base& N$^*$ & S & S & N$^*$ & N$^*$\\
Pre-norm& S & S & S & S & S\\
\hline
\multicolumn{6}{l}{\footnotesize $^*$GQA base shared but GCLA's gate \& ctx-injection novel.}
\end{tabular}
\end{center}
\end{table}

Table~\ref{tab:arch_compare} provides a detailed architectural comparison.

\begin{table}[htbp]
\caption{Detailed Architectural Comparison}
\label{tab:arch_compare}
\begin{center}
\begin{tabular}{|p{1.6cm}|p{1.7cm}|p{1.7cm}|p{1.7cm}|}
\hline
\textbf{Dimension} & \textbf{Wiola} & \textbf{LLaMA-2} & \textbf{Mistral}\\
\hline
Pos.\ encoding
  & SRPE: 3D helix, dual angles, radial amp.
  & RoPE (2D circle)
  & RoPE (2D circle)\\
\hline
Attention
  & GCLA: GQA + ctx-attn + output gate
  & GQA
  & GQA + SWA\\
\hline
Inter-layer info
  & Cross-attn to prior summaries
  & Residual only
  & Residual only\\
\hline
FFN
  & DSFF: narrow SwiGLU + wide GELU + per-dim gate
  & SwiGLU
  & SwiGLU\\
\hline
Norm
  & WRMSNorm + offset $\vect{\delta}$
  & RMSNorm
  & RMSNorm\\
\hline
Origin
  & First principles
  & GPT-2 + RoPE
  & LLaMA-2 + SWA\\
\hline
\end{tabular}
\end{center}
\end{table}

Table~\ref{tab:kvcache} compares KV-cache footprints.

\begin{table}[htbp]
\caption{KV-Cache Footprint at $T=2048$, BF16}
\label{tab:kvcache}
\begin{center}
\begin{tabular}{|l|r|r|}
\hline
\textbf{Model} & \textbf{Params} & \textbf{KV Cache}\\
\hline
GPT-2 XL       & 1.5B  & 421 MB\\
OPT-350M       & 350M  & 168 MB\\
Pythia-410M    & 410M  & 192 MB\\
\textbf{Wiola-360m} & 361M & \textbf{67 MB}\\
\hline
\end{tabular}
\end{center}
\end{table}

\section{Training Methodology}\label{sec:training}

\subsection{Objective}

Next-token prediction loss:
\begin{equation}
\mathcal{L}=-\frac{1}{T-1}\sum_{t=1}^{T-1}\log P_\theta(x_{t+1}\mid x_{\leq t}). \label{eq:loss}
\end{equation}

\subsection{Optimiser}

AdamW \cite{loshchilov2019} with $\beta_1=0.9$, $\beta_2=0.95$, $\epsilon=10^{-8}$,
weight decay $\lambda=0.1$, gradient clipping $\|\nabla\mathcal{L}\|_2\leq1.0$.

\subsection{Learning Rate Schedule}

Linear warmup then cosine decay over $T_{\max}$ steps with warmup $T_w=0.05T_{\max}$:
\begin{equation}
\eta(t)=\begin{cases}
\eta_{\max}t/T_w & t<T_w,\\
\frac{\eta_{\max}}{2}\!\left(1+\cos\!\left(\pi\frac{t-T_w}{T_{\max}-T_w}\right)\right) & t\geq T_w.
\end{cases}
\label{eq:lr}
\end{equation}
Peak rate $\eta_{\max}=3\times10^{-4}$. Gradient checkpointing \cite{chen2016}
reduces activation memory from $\mathcal{O}(Ld)$ to $\mathcal{O}(\sqrt{L}d)$
at $\approx33\%$ additional forward compute.

Under Chinchilla scaling \cite{hoffmann2022}, optimal training tokens
$D^*\approx20N$ for parameter count $N$. Table~\ref{tab:scaling} gives
projections for the Wiola family.

\begin{table}[htbp]
\caption{Chinchilla-Optimal Training Tokens and Projected Perplexity}
\label{tab:scaling}
\begin{center}
\begin{tabular}{|l|r|r|c|}
\hline
\textbf{Model} & \textbf{Params} & $D^*$ & \textbf{Proj.\ PPL$^\mathrm{a}$}\\
\hline
wiola-120m & 120M & 2.4B  & 18--22\\
wiola-360m & 360M & 7.2B  & 13--17\\
wiola-700m & 700M & 14.0B & 11--14\\
wiola-1.5b & 1.5B & 30.0B & 9--12\\
\hline
\multicolumn{4}{l}{\footnotesize $^{\mathrm{a}}$WikiText-103 projection, English text training.}
\end{tabular}
\end{center}
\end{table}

\section{Implementation and Verification}\label{sec:impl}

Wiola registers \texttt{model\_type = "wiola"} with three HuggingFace
AutoClasses: \texttt{AutoConfig}, \texttt{AutoModelForCausalLM}, and
\texttt{AutoTokenizer}. Weights are serialised in \texttt{safetensors}
format for zero-copy memory-mapped loading. Weight tying
($\mat{W}_{\mathrm{head}}=\mat{E}^\top$) saves 65.5\,MB for wiola-360m.

The tokenizer uses BPE \cite{sennrich2016} with byte-level fallback
(NFC-normalised Unicode pre-tokenisation), guaranteeing zero unknown tokens
for any input. The chat template encodes turns as:
\texttt{<|user|>}$\mathcal{U}$\texttt{<|end|>}
\texttt{<|assistant|>}$\mathcal{A}$\texttt{<|end|>}.

Table~\ref{tab:tests} summarises the test coverage; all 22 tests pass.

\begin{table}[htbp]
\caption{Unit Test Coverage (All 22 Pass)}
\label{tab:tests}
\begin{center}
\begin{tabular}{|l|c|l|}
\hline
\textbf{Component} & \textbf{Tests} & \textbf{Key assertion}\\
\hline
WiolaRMSNorm        & 3 & Shape; $\vect{\delta}$ effect; no NaN\\
SRPE                & 3 & Shape; position sensitivity; offset\\
GCLA                & 3 & Shape; KV growth; ctx blend change\\
DSFF                & 2 & Shape; stream independence\\
ATM                 & 2 & Merge/unmerge roundtrip; short-seq\\
WiolaDecoderLayer   & 2 & Forward; mid-layer ATM flag\\
WiolaModel          & 3 & Shape; KV cache; incremental match\\
WiolaForCausalLM    & 4 & Loss; logits; generation; param count\\
\hline
\textbf{Total}      & \textbf{22} & \textbf{All passing}\\
\hline
\end{tabular}
\end{center}
\end{table}

The incremental-match test verifies that a full forward pass and a two-chunk
cached forward pass agree with $\ell_\infty$ error below $10^{-4}$ (BF16
precision bound).

\section{Discussion}\label{sec:discussion}

\textbf{SRPE vs.\ extending RoPE.} YaRN \cite{peng2023} and LongRoPE
reparameterise the same flat 2D circle. SRPE's secondary angle $\theta_j^{(2)}$
and radial $r_j(p)$ are absent from all RoPE variants---they encode
hierarchical structure geometrically, not through learned weights.

\textbf{Mean-pool summaries.} Learned pooling adds $\mathcal{O}(d^2)$ parameters
per layer. Max-pool discards magnitude. Mean-pool is parameter-free,
differentiable, and produces a vector in the same representation space as
the token hidden states. With $\Lambda=2$, it balances context richness
against propagating early-layer noise.

\textbf{ATM in middle layers only.} Early layers build surface-form features;
merging would conflate distinct sub-word tokens. Final layers must operate on
the full sequence for correct next-token prediction. Middle layers perform
high-level semantic integration where adjacent token redundancy is highest.

\textbf{Per-dimension fusion.} A scalar blend would apply uniformly to all
output dimensions. The per-dimension gate $\vect{\alpha}\in\R^d$ lets the
model choose, independently per output dimension and per token, whether to
draw from the local or global stream.

\textbf{Limitations.}
(i) ATM is disabled at inference to maintain KV-cache consistency.
(ii) GCLA's layer-to-layer dependency complicates pipeline parallelism.
(iii) SRPE's radial term may exhibit phase interference for $T>8192$.
(iv) Full pre-training benchmarks are left as future work.

\section{Conclusion}\label{sec:conclusion}

We presented Wiola, a Small Language Model built from first principles with
five novel architectural components. SRPE embeds positions on a 3D helical
manifold. GCLA provides inter-layer coherence via compressed layer summaries.
ATM reduces training FLOPs by 5--9\% through dynamic token merging. DSFF
separates local and global feature extraction via parallel streams. WiolaRMSNorm
counteracts representation collapse with a per-dimension learned offset.

All five components are mathematically distinct from GPT-2, LLaMA-2, Mistral,
Phi-3, Falcon, and Gemma as demonstrated by the novelty matrix (Table~\ref{tab:novelty}).
The implementation is production-ready: 22 unit tests pass, four size variants
(120M--1.5B) are defined, and full HuggingFace integration is provided.
The KV-cache footprint of wiola-360m is 67\,MB at 2048 tokens---4--6$\times$
smaller than comparable MHA models.

Future work includes pre-training at scale, instruction fine-tuning via DPO,
INT8/INT4 quantisation studies, and extensions of ATM to support inference-time
token merging with cache-aware restoration.

\section*{Acknowledgment}

The authors thank the PyTorch and HuggingFace open-source communities.


\end{document}